\documentclass[conference]{IEEEtran}

\usepackage{url}
\usepackage{multirow}
\usepackage[utf8]{inputenc}
\usepackage[T1]{fontenc}
\usepackage{hyperref}
\usepackage{placeins}
\usepackage{subcaption}

\ifCLASSINFOpdf
   \usepackage[pdftex]{graphicx}
   \graphicspath{{figs/}}
   \DeclareGraphicsExtensions{.pdf,.jpeg,.png}
\else
   \usepackage[dvips]{graphicx}
   \graphicspath{{../figs/}}
   \DeclareGraphicsExtensions{.eps}
\fi

\makeatletter
\def\blfootnote{\xdef\@thefnmark{}\@footnotetext}
\makeatother

\begin{document}
%
\title{Are pre-trained CNNs good feature extractors for anomaly detection in surveillance videos?}

\newif\iffinal
\finaltrue
\newcommand{\jemsid}{33}

\iffinal
  \author{%
    \IEEEauthorblockN{Tiago S. Nazare$^{*\ddag}$, Rodrigo F. de Mello$^{*}$, Moacir A. Ponti$^{*}$}
    \IEEEauthorblockA{%
      $^{*}$ICMC, University of S\~ao Paulo, S\~ao Carlos, Brazil.\\
      $^{\ddag}$Data Science Team, Ita\'u-Unibanco, S\~ao Paulo, Brazil.\\
      Email: tiagosn@usp.br, mello@icmc.usp.br, moacir@icmc.usp.br} 
  }
\else
  \author{WVC paper ID: \jemsid \\ }
\fi

\maketitle

\begin{abstract}
Recently, several techniques have been explored to detect unusual behaviour in surveillance videos. Nevertheless, few studies leverage features from pre-trained CNNs and none of then present a comparison of features generate by different models. Motivated by this gap, we compare features extracted by four state-of-the-art image classification networks as a way of describing patches from security video frames. We carry out experiments on the Ped1 and Ped2 datasets and analyze the usage of different feature normalization techniques. Our results indicate that choosing the appropriate normalization is crucial to improve the anomaly detection performance when working with CNN features. Also, in the Ped2 dataset our approach was able to obtain results comparable to the ones of several state-of-the-art methods. Lastly, as our method only considers the appearance of each frame, we believe that it can be combined with approaches that focus on motion patterns to further improve performance.


\end{abstract}

\IEEEpeerreviewmaketitle

\section{Introduction}

\blfootnote{Any opinions, findings, and conclusions expressed in this manuscript are those of the authors and do not necessarily reflect the views, official policy or position of the Ita\'u-Unibanco, FAPESP and CNPq.}

Nowadays security cameras are widely employed to monitor public spaces -- such as malls, squares and universities. Yet, those surveillance cameras may be ineffective, mainly because each video feed needs a person constantly watching it. Keeping track of the events in security videos is very hard for humans, especially for two reasons: 1) a single person is responsible for monitoring several cameras simultaneously~\cite{Dee2008}; 2) it is hard for people to maintain an acceptable level of attention when watching this kind of video~\cite{Haering2008}. Due to the aforementioned issues, security footage is of little help with regards to preventing dangerous situations and end up being used mostly for investigation purposes, after something already happened.

The ineffectuality of surveillance systems motivated the machine vision community to work on automated systems to detect unusual behaviour in security videos, consequently, we have seen outstanding advances in this area. Over the last few years a great deal of methods have been proposed to detect anomalies in videos. Among the approaches employed to tackle this issue we have: time series decomposition of optical-flow~\cite{Ponti2017}, optical-flow features~\cite{Colque2015, Adam2008}, dictionary learning~\cite{Li2015}, auto-encoders~\cite{Xu2015} and GANs (generative adversarial networks)~\cite{Ravanbakhsh2017}. Despite all progress obtained over the last few years, automatically detecting unusual events in videos remains an open research area and several approaches are yet to be explored.

One kind of approach that -- to the best of our knowledge -- have not yet been broadly investigated in anomaly detection scenarios is transfer-learning. This technique consists of leveraging knowledge obtained from solving a particular problem to solve a different one (on a related domain). With regards to CNNs most transfer learning applications use pre-trained networks as feature extractors or as starting point for training. In recent years, such usage of pre-trained networks has been considered to be very effective by various studies~\cite{Ponti2017b}, even when the original and target domains are considerably different~\cite{Razavian2014}. Still, such performance is not guaranteed for every scenario, as pointed out by some other studies~\cite{Dodge16, Nazare2017}. 

Motivated by the aforementioned results and the lack of investigation regarding the usage of pre-trained CNNs to detect unusual behaviour in security videos, in this paper, we devote our efforts to evaluate the application of several state-of-the-art image classification CNNs as features extractor machines for surveillance footage. We compared features generated by four networks (VGG-16~\cite{Simonyan2014}, RestNet-50~\cite{He2016}, Xception~\cite{Chollet2017} and DenseNet-121~\cite{Huang2017}) as a way to describe appearance for frame regions of security videos. Despite neglecting the motion part of anomaly detection in videos, our experiments (conducted using the Ped1 and Ped2 UCSD datasets~\cite{Mahadevan2010, Li2014}) show that those features are able to achieve competitive results.

\section{Related work}

\begin{figure*}[!ht]
\centering
\includegraphics[width=6.5in]{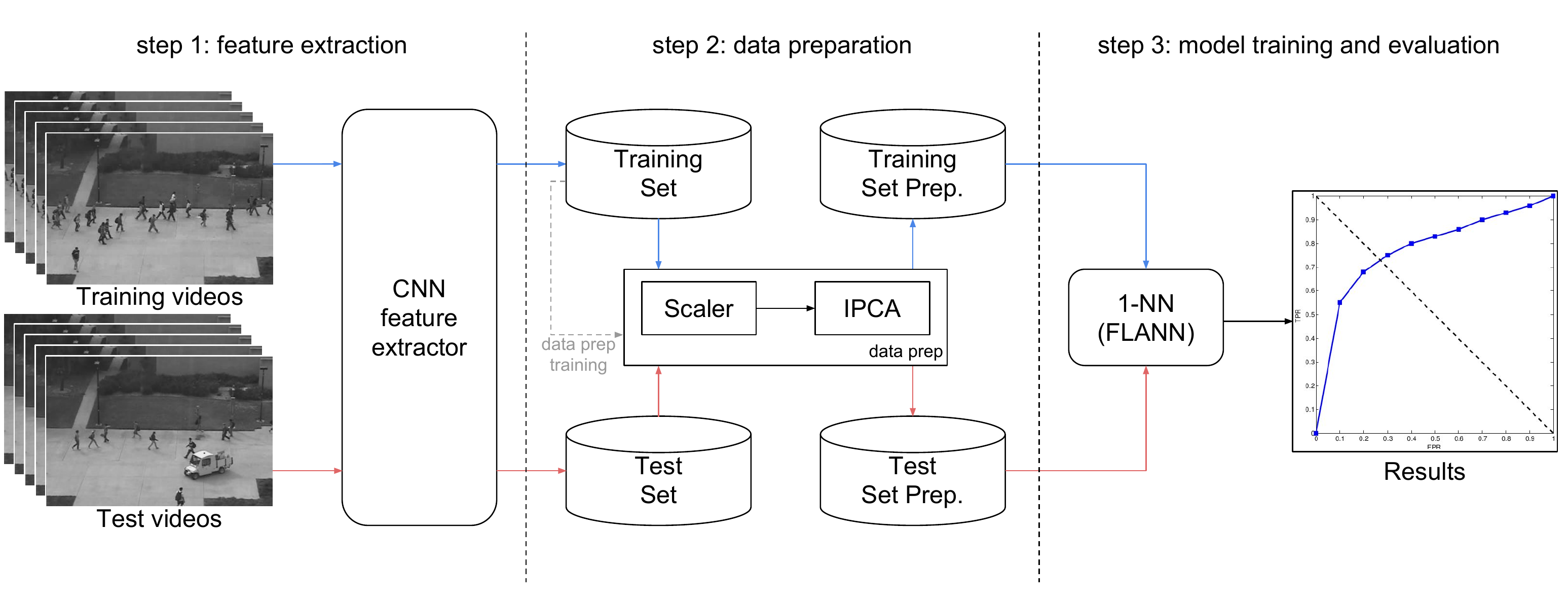}
\caption{Experimental setup diagram.}
\label{fig:experimental_setup}
\end{figure*}

Recently, the usage of CNN features has been explored to detect unusual activities in surveillance footage. In~\cite{Ravanbakhsh2018}, the authors describe image regions using AlexNet~\cite{Krizhevsky2012} and, then, track variations in such description in order to detect anomalies. With this tracking, they are able to use a image based CNN to find both motion and appearance anomalies. 

In~\cite{Sultani2018} a pre-trained C3D model~\cite{Tran2015} was employed for feature extraction. Such model was originally designed for action recognition in videos, thus it learns spatio-temporal representations, which is very useful to deal with both motion and appearance abnormalities. Nevertheless, in~\cite{Sultani2018} the pre-trained model is used in a classification setup, therefore it has access to instances of both normal and anomalous behaviour during training.

In spite of the fact that are some studies where pre-trained networks were employed to detect anomalies in security videos, they do not compare several models in order to determine which one is best suited for the application. For this reason, in this paper we design a experimental setup and present several experimental results --- employing different network architectures and data normalization procedures --- in order to shed some light on this issue. Instead of attempting to outperform state-of-the-art methods, we aim at a better understanding of how off-the-shelf CNN features behave for the application of video surveillance, providing a guideline for future research on how to choose the appropriate model for this task.

\section{Experimental setup}
\label{sec:experimental_setup}

In order to measure how well the features extracted from pre-trained image classification CNNs perform, when detecting anomalies in surveillance videos, we employed the experimental setup depicted in Figure~\ref{fig:experimental_setup}. This experimental setup has three steps.

In the \textbf{first step} of our setup we start by converting each video frame to $384 \times 256$ pixels. Next, we take image regions of $32 \times 32$ pixels using a stride of 16 pixels and perform a forward pass of each of these patches through the convolutional part of a CNN pre-trained on the ImageNet dataset~\cite{imagenet09}. This image patch end up being represented by $d$ features, where the value of $d$ depends on the network -- it is equal to the number of filters (neurons) at the last convolutional layer of the network. At this point one can notice that using the convolutional part of the network -- instead of the entire network -- is rather convenient in our framework. This is due to the fact that when using the entire networks we have to provide input images of the same size of the original dataset (e.g. $224\times 224$), while with the convolutional part of the network we do not have such constraint. In this part of our setup, we investigated features generated by the following state-of-the-art image classification models (trained on ImageNet): VGG-16~\cite{Simonyan2014}, ResNet-50~\cite{He2016}, Xception~\cite{Chollet2017} and DenseNet-121~\cite{Huang2017}. Please refer to Section~\ref{sec:cnns} for more details on each model and also for the number of features generated by each one of them.

In the \textbf{second step} we prepared the data obtained on the first step in order to use it with an anomaly detection method. To do so, we started by normalizing the data, in this part we tested the z-score, 0-1, L1 and L2 normalization methods (see Section~\ref{sec:data_norm} for a description of these techniques). Given that we employed the nearest neighbour technique to detect anomalies, this normalization step is fundamental to the performance of the system. After normalizing the data we used the Incremental PCA (IPCA) algorithm~\cite{Ross2008} to reduce data dimensionality to 50 and 100 dimensions. This particular method was chosen for two reasons: 1) it can be used to reduce the number of features and, consequently, speed up the computation of nearest neighbours; 2) it does not need to load the entire dataset into RAM in order to compute its transformation, so it can deal with large datasets such as those used in this paper. 

Lastly, in the \textbf{third step} of our experimental setup, we train an 1-NN model using the datasets obtained in step two (with 50 or 100 features depending on the experiment). Then, we find the distance of all the test set instances to their nearest neighbour in the training set. Those distances are used as an anomaly score. Hence, the higher the distance, the more a instance (image region) is considered to be anomalous. It is important to notice that we train only one model to work with descriptions from the entire image. Also, aiming to speed up our anomaly detection framework, we used the approximate nearest neighbour method from~\cite{Muja2014}. 
Next, we obtain the anomaly score of a frame simply by taking the max score among its patches. Based on those frame scores we can compute the AUC (Area Under ROC Curve) and EER (Equal Error Ratio) on a frame-level classification.

Our experiments were carried out on the UCSD video anomaly detection datasets: Ped1 and Ped2~\cite{Mahadevan2010, Li2014}. More information about the two datasets are presented Section~\ref{sec:datasets}.

\subsection{Data normalization}
\label{sec:data_norm}

In our experiments we used the following four data normalization methods:

\begin{itemize}
    \item \textit{z-score}: outputs features centered at the origin (i.e. with zero mean) and with unit standard deviation by computing:

    $$ f_i = \frac{f_i - mean(f)}{std(f)}, $$
    
    \noindent where $f_i$ is the value of feature $f$ for instance $i$, and the mean and standard deviation of $f$ is obtained using the entire training set;
    
    \item \textit{0-1}: outputs features in the $[0, 1]$ interval. To do so, it uses the following formula:
    
    $$ f_i = \frac{f_i - min(f)}{max(f) - min(f)}, $$
    
    \noindent where $f_i$ is the value of feature $f$ for instance $i$ and the max and min of $f$ are obtained from the training set;
    
    \item \textit{L1}: in this method all instances (dataset rows) are scaled so their L1 norm are equal to one, as follows:
    
    $$ x = \frac{x}{\sum_f |x_f|}, $$
    
    \noindent where $x$ is an instance and $|x_f|$ is the absolute value of the \textit{f-th} feature of $x$.
    
    \item \textit{L2}: similar to the previous one, but using the L2 norm instead, as in the equation bellow: 
    
    $$ x = \frac{x}{\sqrt{\sum_f x^{2}_{f}}}. $$
\end{itemize}





\subsection{CNN models}
\label{sec:cnns}

Bellow, we quickly present some of the main characteristics of the four CNNs used as feature extractors in our experiments.

\paragraph*{VGG-16~\cite{Simonyan2014}} is composed of 16 layers, 13 of them convolutional and the remaining 3 dense layers. All convolutional layers use $3 \times 3$ kernels and ReLU as activation function. Also, those 13 layers are divided into 6 groups and each group has a max-pooling layer at its end.

\begin{figure}[!ht]
\centering
\includegraphics[width=1.5in]{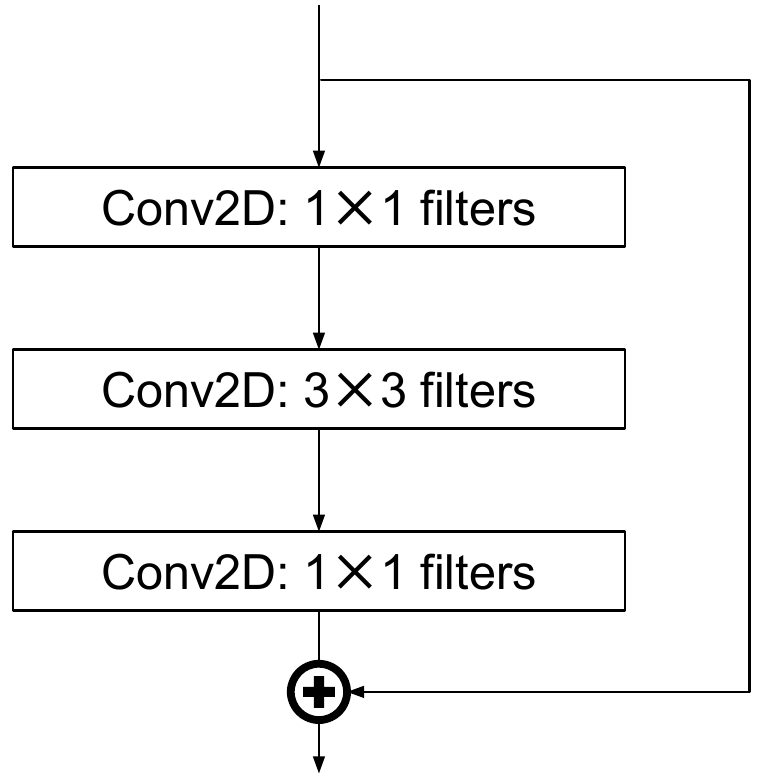}
\caption{Residual unit employed by ResNet-50~\cite{He2016}.}
\label{fig:residual_unit}
\end{figure}

\paragraph*{ResNet-50~\cite{He2016}} uses an alternative layer configuration method, called residual units. Residual units -- such as the one depicted in Figure~\ref{fig:residual_unit} -- allow deeper models to be learned by using shortcut connections to avoid vanishing/exploding gradients during training. This network has 49 convolutional layers and only one fully connected layer. All but one of its convolutional layers are organized into 16 residual units like the one of Figure~\ref{fig:residual_unit}.

\begin{figure}[!ht]
\centering
\includegraphics[width=3.3in]{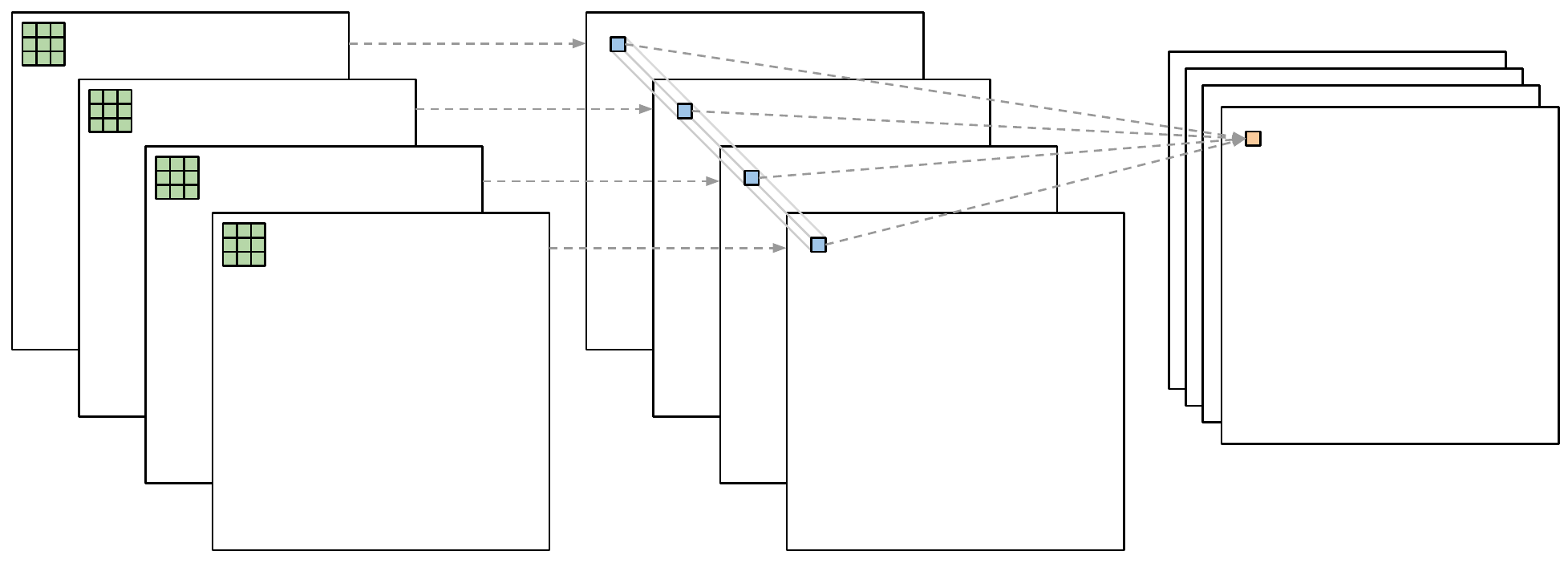}
\caption{Depthwise separable convolution layer used by Xception~\cite{Chollet2017}.}
\label{fig:separable_conv}
\end{figure}

\paragraph*{Xception~\cite{Chollet2017}} is inspired by Inception V3~\cite{Szegedy2016}, and built by replacing the inception modules with depthwise separable convolutions. This type of convolution is performed by, first, applying convolutions (e.g. using $3 \times 3$ kernels) to each tensor channel separately and, then, applying an $1 \times 1$ convolution to all channels, this process is illustrated in Figure~\ref{fig:separable_conv}. By using this type of operation the Xception model was able to outperform Inception V3 for image classification using the same number of trainable parameters.

\paragraph*{DenseNet-121~\cite{Huang2017}} is an architecture similar to ResNets because they also use shortcut/skipping connections. Nevertheless, instead of employing this type of connection in some specific parts of the network, they use shortcut connections to connect every layer to all of its subsequent layers. The authors claim this approach avoids vanishing/exploding gradients and allows the training of even deeper networks when compared to ResNets.

As previously mentioned, those models have different number of trainable parameters and generate descriptors of different sizes. In Table~\ref{tab:trainable_parameters}, we present the number of trainable parameters in the convolutional part (feature extractor part) and the number of generated features of each CNN used in our experiments.

\begin{table}[!ht]
\centering
\caption{Number of trainable parameters and output features for each one of the network architectures used in our experiments.}
\label{tab:trainable_parameters}
\begin{tabular}{|l|c|c|}
\hline
\multicolumn{1}{|c|}{\textbf{feature extractor}} & \textbf{\# of trainable parameters} & \textbf{\# of features} \\ \hline
VGG-16~\cite{Simonyan2014}                                           & 14,714,688                          & 512                     \\ \hline
ResNet-50~\cite{He2016}                                        & 23,534,592                          & 2048                    \\ \hline
Xception~\cite{Chollet2017}                                         & 20,806,952                          & 2048                    \\ \hline
DenseNet-121~\cite{Huang2017}                                     & 6,953,856                           & 1024                    \\ \hline
\end{tabular}
\end{table}

\subsection{Datasets}
\label{sec:datasets}

\begin{figure*}[!ht]
\centering
\begin{subfigure}[b]{0.33\textwidth}
        \includegraphics[width=\textwidth]{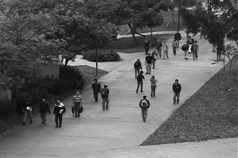}
\end{subfigure}%
~ 
\begin{subfigure}[b]{0.33\textwidth}
        \includegraphics[width=\textwidth]{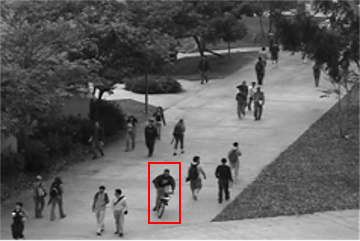}
\end{subfigure}%
~ 
\begin{subfigure}[b]{0.33\textwidth}
        \includegraphics[width=\textwidth]{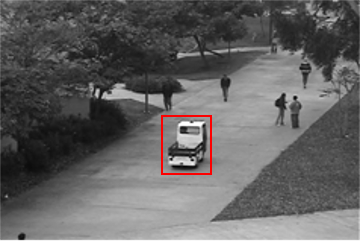}
\end{subfigure}

~

\begin{subfigure}[b]{0.33\textwidth}
        \includegraphics[width=\textwidth]{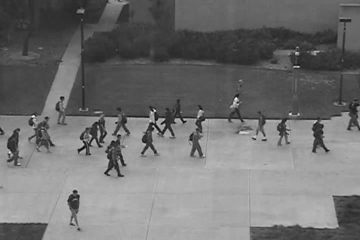}
\end{subfigure}%
~ 
\begin{subfigure}[b]{0.33\textwidth}
        \includegraphics[width=\textwidth]{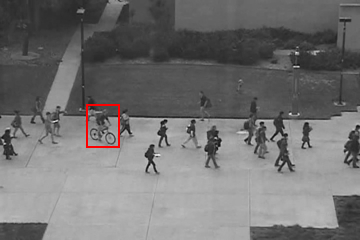}
\end{subfigure}%
~ 
\begin{subfigure}[b]{0.33\textwidth}
        \includegraphics[width=\textwidth]{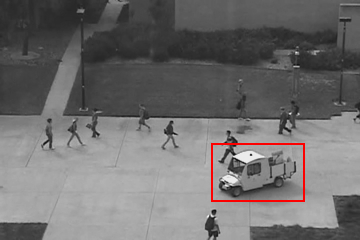}
\end{subfigure}
\caption{Frames from Ped1 (first row) and Ped2 (second row), with anomalies manually highlighted by red boxes.}
\label{fig:ucsd_frames}
\end{figure*}

Both datasets employed in our experiments were obtained for stationary surveillance cameras at the UCSD (University of California, San Diego) campus~\cite{Mahadevan2010, Li2014}. This means that all of their videos are real (they were not staged). Each dataset contains videos from a single camera and only the test set has anomalous events. Some specifications and sample images from both datasets are presented in Table~\ref{ucsd_specs} and Figure~\ref{fig:ucsd_frames}, respectively.

\begin{table}[!ht]
\centering
\caption{Video specifications for Ped1 and Ped2.}
\label{ucsd_specs}
\begin{tabular}{|c|c|c|}
\hline
\multirow{2}{*}{\textbf{characteristic}} & \multicolumn{2}{c|}{\textbf{dataset}} \\ \cline{2-3} 
                               & \textbf{Ped1}     & \textbf{Ped2}     \\ \hline
training videos            & 34                & 16                \\ \hline
test videos             & 36                & 12                \\ \hline
frame resolution              & $238 \times 158$  & $360 \times 240$  \\ \hline
fps                     & 10                & 10                \\ \hline
frames per video        & 200               & $120-180$        \\ \hline
\end{tabular}
\end{table}

\subsection{Reproducibility remarks}

In our experiments we used the \texttt{Keras}~\cite{keras} library, and its pre-trained models, for feature extraction. In order to normalize our features and compute IPCA, we used the \texttt{Scikit-learn}~\cite{sklearn} library. Lastly, for the approximate nearest neighbour distance computation we have used the \texttt{FLANN}~\cite{flann} library. To facilitate the reproduction of our results our source code is publicly available\footnote{
    \iffinal
    Our source code is available at \url{https://github.com/tiagosn/cnn_features_anomaly_detection}.
    \else
    \textbf{For anonymity purposes this url was omitted.}
    \fi
}.

\section{Results and Discussion}

As stated in Section~\ref{sec:experimental_setup}, we trained nearest neighbour models using features extracted from four pre-trained CNNs. In those tests, before training our nearest neighbour models, we first normalized our features and, then, used IPCA to reduce the number of dimensions. Aiming at obtaining the best results out of our features we experimented with several configuration regarding both the normalization technique and the number of dimensions used after the IPCA transformation. The results of such experiments are presented in Table~\ref{tab:results_cnns_ucsd}.

\begin{table*}[!ht]
\caption{Frame level detection results on Ped1 and Ped2 datasets using several pre-trained CNNs as feature extractors, various feature normalization techniques and approximate nearest neighbour. Please keep in mind that the lower the EER, the better and the higher the AUC, the better.}
\label{tab:results_cnns_ucsd}
\centering
\begin{tabular}{|c|c|c|c|l|l|l|}
\hline
\multicolumn{1}{|l|}{\multirow{3}{*}{\textbf{feature extractor}}} & \multicolumn{1}{l|}{\multirow{3}{*}{\textbf{IPCA dimensions}}} & \multicolumn{1}{l|}{\multirow{3}{*}{\textbf{normalization method}}} & \multicolumn{4}{c|}{\textbf{dataset}}                                                                                        \\ \cline{4-7} 
\multicolumn{1}{|l|}{}                                            & \multicolumn{1}{l|}{}                                          & \multicolumn{1}{l|}{}                                               & \multicolumn{2}{c|}{\textbf{Ped1}}                   & \multicolumn{2}{c|}{\textbf{Ped2}}                                    \\ \cline{4-7} 
\multicolumn{1}{|l|}{}                                            & \multicolumn{1}{l|}{}                                          & \multicolumn{1}{l|}{}                                               & \textbf{AUC}     & \multicolumn{1}{c|}{\textbf{EER}} & \multicolumn{1}{c|}{\textbf{AUC}} & \multicolumn{1}{c|}{\textbf{EER}} \\ \hline
\multirow{8}{*}{VGG-16~\cite{Simonyan2014}}                                           & \multirow{4}{*}{50}                                            & 0-1                                                           & 63.98\%          & 40.85\%                           & 63.81\%                           & 42.90\%                           \\ \cline{3-7} 
                                                                  &                                                                & z-score                                                                 & 63.35\%          & 42.02\%                           & 64.98\%                           & 40.05\%                           \\ \cline{3-7} 
                                                                  &                                                                & L1                                                                  & 59.01\%          & 43.28\%                           & 63.73\%                           & 39.95\%                           \\ \cline{3-7} 
                                                                  &                                                                & L2                                                                  & 59.16\%          & 43.15\%                           & 63.37\%                           & 40.13\%                           \\ \cline{2-7} 
                                                                  & \multirow{4}{*}{100}                                           & 0-1                                                           & \textbf{64.06\%}          & 41.02\%                           & 63.42\%                           & 40.29\%                           \\ \cline{3-7} 
                                                                  &                                                                & z-score                                                                 & 63.78\%          & 41.13\%                           & 64.97\%                           & 38.53\%                           \\ \cline{3-7} 
                                                                  &                                                                & L1                                                                  & 63.62\%          & \textbf{40.40\%}                           & 65.49\%                           & 38.40\%                           \\ \cline{3-7} 
                                                                  &                                                                & L2                                                                  & 61.02\%          & 42.72\%                           & 62.84\%                           & 40.70\%                           \\ \hline
\multirow{8}{*}{ResNet-50~\cite{He2016}}                                        & \multirow{4}{*}{50}                                            & 0-1                                                           & 59.10\%          & 44.94\%                           & 62.05\%                           & 42.27\%                           \\ \cline{3-7} 
                                                                  &                                                                & z-score                                                                 & 60.95\%          & 43.17\%                           & 79.59\%                           & 28.88\%                           \\ \cline{3-7} 
                                                                  &                                                                & L1                                                                  & 55.26\%          & 45.73\%                           & 45.59\%                           & 51.93\%                           \\ \cline{3-7} 
                                                                  &                                                                & L2                                                                  & 63.60\%          & 41.25\%                           & 71.05\%                           & 34.53\%                           \\ \cline{2-7} 
                                                                  & \multirow{4}{*}{100}                                           & 0-1                                                           & 59.65\%          & 44.66\%                           & 69.02\%                           & 37.08\%                           \\ \cline{3-7} 
                                                                  &                                                                & z-score                                                                 & 61.98\%          & 42.21\%                           & 83.90\%                           & 23.33\%                           \\ \cline{3-7} 
                                                                  &                                                                & L1                                                                  & 55.48\%          & 45.93\%                           & 47.49\%                           & 51.46\%                           \\ \cline{3-7} 
                                                                  &                                                                & L2                                                                  & 62.67\%          & 42.19\%                           & 70.08\%                           & 34.89\%                           \\ \hline
\multirow{8}{*}{Xception~\cite{Chollet2017}}                                         & \multirow{4}{*}{50}                                            & 0-1                                                           & 59.95\%          & 44.18\%                           & 86.61\%                           & 21.82\%                           \\ \cline{3-7} 
                                                                  &                                                                & z-score                                                                 & 59.16\%          & 44.23\%                           & 87.33\%                           & 21.66\%                           \\ \cline{3-7} 
                                                                  &                                                                & L1                                                                  & 51.59\%          & 48.91\%                           & 63.72\%                           & 41.32\%                           \\ \cline{3-7} 
                                                                  &                                                                & L2                                                                  & 60.02\%          & 44.21\%                           & 80.78\%                           & 25.39\%                           \\ \cline{2-7} 
                                                                  & \multirow{4}{*}{100}                                           & 0-1                                                           & 60.68\%          & 43.75\%                           & 87.94\%                           & \textbf{19.55\%}                           \\ \cline{3-7} 
                                                                  &                                                                & z-score                                                                 & 59.61\% & 43.59\%                           & \textbf{88.93\%}                           & 20.02\%                           \\ \cline{3-7} 
                                                                  &                                                                & L1                                                                  & 51.99\%          & 48.71\%                           & 65.21\%                           & 40.35\%                           \\ \cline{3-7} 
                                                                  &                                                                & L2                                                                  & 58.83\%          & 43.56\%                           & 82.03\%                           & 24.58\%                           \\ \hline
\multirow{8}{*}{DenseNet-121~\cite{Huang2017}}                                     & \multirow{4}{*}{50}                                            & 0-1                                                           & 63.65\%          & 41.02\%                           & 82.91\%                           & 23.58\%                           \\ \cline{3-7} 
                                                                  &                                                                & z-score                                                                 & 62.04\%          & 41.83\%                           & 72.57\%                           & 34.81\%                           \\ \cline{3-7} 
                                                                  &                                                                & L1                                                                  & 63.00\%          & 40.60\%                           & 73.73\%                           & 31.98\%                           \\ \cline{3-7} 
                                                                  &                                                                & L2                                                                  & 63.15\%          & 41.34\%                           & 72.79\%                           & 32.87\%                           \\ \cline{2-7} 
                                                                  & \multirow{4}{*}{100}                                           & 0-1                                                           & 63.16\%          & 41.88\%                           & 84.61\%                           & 23.06\%                           \\ \cline{3-7} 
                                                                  &                                                                & z-score                                                                 & 62.57\%          & 41.44\%                           & 83.07\%                           & 24.31\%                           \\ \cline{3-7} 
                                                                  &                                                                & L1                                                                  & 62.73\%          & 41.40\%                           & 78.09\%                           & 28.16\%                           \\ \cline{3-7} 
                                                                  &                                                                & L2                                                                  & 62.71\%          & 42.13\%                           & 78.05\%                           & 27.07\%                           \\ \hline
\end{tabular}
\end{table*}

By looking at those results it is possible to notice that the normalization method can greatly impact anomaly detection results. Also, it is possible to realize that not all networks have the same behaviour with regards to normalization. With ResNet-50 and Xcpetion features we usually obtained the best results by using the z-score normalization, while with DenseNet-121 the best results were obtained by using the 0-1 normalization. Considering the number of dimensions used to train the model, in most cases, going from 50 to 100 features has shown to be beneficial. Nevertheless, it is important to keep in mind that increasing the number of features makes the nearest neighbour inference slower.

Now, we compare the best results obtained in our test against classic and state-of-the-art results methods on the UCSD (Ped1 and Ped2) datasets. Such comparison is presented in Table~\ref{tab:results_comparison}. Although the CNN features are not better when compared to state-of-the-art results, the performance was reasonable. Specially, considering that feature extraction was not trained with any data from the target task (i.e. Ped1 and Ped2 frames) and motion information was ignored when we only described individual frames. In particular for Ped2, the results are comparable to the ones obtained by some state-of-the-art methods, therefore they can be considered as a good baseline for this dataset. In the case of Ped1, the results are only comparable to some classic methods. We believe that one of the main reasons for these lower performance is the fact that Ped1 has changes in perspective -- objects change size (regarding number of pixels) according to the image region that they are in. Such characteristic can hamper the performance in a setup like ours, where a single model is trained to deal with image patches from the entire frame. 

\begin{table}[]
\centering
\caption{Comparison of frame-level anomaly detection on Ped1 and Ped2 datasets. Please note that the lower the EER, the better and the higher the AUC, the better.}
\label{tab:results_comparison}
\begin{tabular}{|c|c|c|c|c|}
\hline
\multirow{3}{*}{\textbf{method}}                               & \multicolumn{4}{c|}{\textbf{dataset}}                                   \\ \cline{2-5} 
                                                               & \multicolumn{2}{c|}{\textbf{Ped1}} & \multicolumn{2}{c|}{\textbf{Ped2}} \\ \cline{2-5} 
                                                               & \textbf{AUC}     & \textbf{EER}    & \textbf{AUC}     & \textbf{EER}    \\ \hline
LMH~\cite{Adam2008}                              & 63.40\% & 38.90\% & 58.10\% & 45.80\%         \\ \hline
MPPCA~\cite{Kim2009}                             & 59.00\% & 40.00\% & 69.30\% & 30.00\%         \\ \hline
Social force~\cite{Mehran2009}                   & 67.50\% & 31.00\% & 55.60\% & 42.00\%         \\ \hline
Sparse reconstruction~\cite{Cong2011}            & -       & 19.00\% & -       & -               \\ \hline
LSA~\cite{Saligrama2012}                         & 92.70\% & 16.00\% & -       & -               \\ \hline
Sparse combination~\cite{Lu2013}                 & 91.80\% & 15.00\% & -       & -               \\ \hline
MDT~\cite{Li2014}                                & 81.80\% & 25.00\% & 82.90\% & 25.00\%         \\ \hline
LNND~\cite{Hu14}                                 & -       & 27.90\% & -       & 23.70\%         \\ \hline
Motion influence map~\cite{Lee2015}              & -       & 24.10\% & -       & \textbf{9.80\%} \\ \hline
Composition pattern~\cite{Li2015}                & -       & 21.00\% & -       & 20.00\%         \\ \hline
HOFM~\cite{Colque2015}                           & 71.50\% & 33.30\% & 89.90\% & 19.00\%         \\ \hline
AMDN~\cite{Xu2015}                               & 92.10\% & 16.00\% & 90.80\% & 17.00\%         \\ \hline
Flow decomposition~\cite{Ponti2017}              & -       & -       & -       & 31.70\%         \\ \hline
Adversarial discriminator~\cite{Ravanbakhsh2017} & \textbf{96.80\%} & \textbf{7.00\%} & \textbf{95.50\%} & 11.00\% \\ \hline
Plug-and-play CNN~\cite{Ravanbakhsh2018}         & 95.70\% & 8.00\%  & 88.40\% & 18.00\%         \\ \hline
\textbf{CNN features (best)}                     & 64.06\% & 40.40\% & 88.93\% & 19.55\%         \\ \hline
\end{tabular}
\end{table}

\section{Conclusion}

In this paper we investigated the usage of pre-trained image classification CNNs as feature extractors to tackle the detection of unusual events on videos. According to our experiments such networks -- when paired with suitable data normalization techniques -- can be very useful to detect abnormal events in security videos. This can be noticed in the Ped2 experiments, where they were able to achieve results comparable to the ones of the state-of-the-art techniques. Another important fact regarding our experiments is that we only explored those features to model appearance anomalies; hence, we neglected the motion part of the anomalies. That being said, we strongly believe that the method proposed in this work can be combined with methods that focus on motion anomalies (e.g. ~\cite{Colque2015, Ponti2017}) to obtain better results.

\section{Future work}

The main points that we intent to investigate in feature research are:

\begin{itemize}
    \item The performance of such features in RGB surveillance videos like the ones of the Avenue dataset~\cite{Lu2013}. Such analysis can help to better understand if gray scale image can hamper the description capabilities of CNNs trained in RGB domains;
    \item Combine our method with some methods that only tackle motion related anomalies, such as the ones presented in~\cite{Colque2015, Ponti2017}, to see if their fusion can further improve performance;
    \item Test the usage of other anomaly detection techniques, for instance Isolation Forest~\cite{Liu2012} with those CNN features.
\end{itemize}

\section*{Acknowledgment}

\iffinal
This work was supported by FAPESP (grants \#2015/04883-0 and \#2016/16111-4), CNPq (grant \#307973/2017-4), Itaú-Unibanco and partially supported by CEPID-CeMEAI (FAPESP grant \#2013/07375-0).
\else
\textbf{For anonymity purposes this section was omitted.}
\fi



%
\bibliographystyle{IEEEtran}
\bibliography{ref}

\end{document}